# Weakly Supervised 3D Classification of Chest CT using Aggregated Multi-Resolution Deep Segmentation Features

Anindo Saha*,a, Fakrul I. Tushar*,b, Khrystyna Farynaa, Vincent M. D'Anniballeb,
Rui Houb,c, Maciej A. Mazurowskib,c, Geoffrey D. Rubinb, Joseph Y. Lob,c

[a] Erasmus+ Joint Master in Medical Imaging and Applications, Universitat de Girona, Spain;
Université de Bourgogne, France; Università degli studi di Cassino, Italy
[b] Department of Radiology, Duke University School of Medicine, Durham, NC
[c] Department of Electrical and Computer Engineering, Duke University, Durham, NC
*Authors with equal contribution to this work.

## ABSTRACT

Weakly supervised disease classification of CT imaging suffers from poor localization owing to case-level annotations, where even a positive scan can hold hundreds to thousands of negative slices along multiple planes. Furthermore, although deep learning segmentation and classification models extract distinctly unique combinations of anatomical features from the same target class(es), they are typically seen as two independent processes in a computer-aided diagnosis (CAD) pipeline, with little to no feature reuse. In this research, we propose a medical classifier that leverages the semantic structural concepts learned via multi-resolution segmentation feature maps, to guide weakly supervised 3D classification of chest CT volumes. Additionally, a comparative analysis is drawn across two different types of feature aggregation to explore the vast possibilities surrounding feature fusion. Using a dataset of 1593 scans labeled on a case-level basis via rule-based model, we train a dual-stage convolutional neural network (CNN) to perform organ segmentation and binary classification of four representative diseases (emphysema, pneumonia/atelectasis, mass and nodules) in lungs. The baseline model, with separate stages for segmentation and classification, results in AUC of 0.791. Using identical hyperparameters, the connected architecture using static and dynamic feature aggregation improves performance to AUC of 0.832 and 0.851, respectively. This study advances the field in two key ways. First, case-level report data is used to weakly supervise a 3D CT classifier of multiple, simultaneous diseases for an organ. Second, segmentation and classification models are connected with two different feature aggregation strategies to enhance the classification performance.

Keywords: CT, convolutional neural network, weak supervision, feature aggregation, 3D classification, lungs

## 1. INTRODUCTION

With over 80 million scans performed annually in the U.S., computed tomography (CT) has become a powerful medical diagnostic tool that provides an extensive 3D overview of a patient's internal anatomy. Although the vast number of slices constituting each scan can pose an overwhelming burden to radiologists, they can greatly benefit machine learning algorithms in their generalization capabilities. As a result, deep neural networks are now at the forefront of automated disease detection and diagnosis technology in medical imaging. While most existing methods implement a simplified 2D approach to classify CT volumes, they fail to utilize the additional spatial information between slices. Meanwhile, in the absence of complete annotation, a 3D approach suffers from the heavy imbalance of noisy labeled data. Thus, following our study on the prospects of a 2D model [1], in this research we propose an alternate solution in the form of a 3D medical classifier, that leverages segmentation features to guide weakly supervised binary (healthy/diseased) classification of chest CT volumes.

Similar state-of-the-art approaches in the domain include Y-Net [2] from S. Mehta et al., proposing an extended 2D segmentation network with an auxiliary classification branch, and the modified M-Net [3] from K. Wong et al., linking individual segmentation and classification models together. While the former introduces joint functionality using a singular stream of shared features, the latter pulls segmentation features from the model decoder and demonstrates their feasibility as a viable basis for curriculum learning. In this research, we explore an alternative approach by extracting all segmentation feature volumes directly from the model backbone and reviewing their corresponding activations for different disease classes. By targeting the most relevant feature maps and exploring two different types of feature aggregation, we highlight the impact of utilizing these fused features in deep neural networks.

## 2. METHODOLOGY

### 2.1 Dataset

The complete classification dataset comprises of 1593 chest and chest-abdominal-pelvis CT scans selected randomly from over 26,000 scans conducted at Duke University Health System between January-April, 2017. Using their radiology reports, these scans have been labelled (as shown in Table 1) on a case-level basis with a pre-trained rule-based model [4,5], to provide weak supervision throughout the training stage. Normal to diseased lungs cases hold a ratio of 1:1.75, reflecting the actual prevalence (64.6%) of diseased cases in the hospital system. The diseased category comprises of four representative diseases (pneumonia/atelectasis, mass, nodules and emphysema), with 38 instances of multiple diseases in the same scan. Prior to training, all scans are resampled into voxels of size $2\times2\times2$ mm$^3$ via B-spline interpolation, clipped to intensity range (-1000, 800) HU, padded to size $112^3$ and normalized to ensure spatial uniformity across the dataset.

Table 1.  Distribution of Multi-Disease Chest CT Dataset

| Diseased Cases | | | | Normal Cases |
|---|---|---|---|---|
| Pneumonia-Atelectasis | Mass | Emphysema | Nodules | |
| **286** | **243** | **279** | **246** | **577** |

Additionally, 50 CT volumes derived from the 4D Extended Cardiac Torso (XCAT) phantom, developed by Duke University Medical Center for multimodality chest-abdominal pelvis imaging research, are used for pre-training the segmentation sub-model [6]. The volumes are accompanied by their segmentation ground truth for 29 classes. 30 diseased scans (exclusive from the classification dataset) are also used to fine-tune this model further, in order to train the anatomical variations that can result from lung abnormalities.

### 2.2 Network Architecture

To assess the effect of aggregated features across multiple diseases, we developed a single model that treated all 4 diseases as a single positive class at train-time. The proposed architecture is a connected network of two sub-models, as shown in Figure 1. The first sub-model is a standard DenseVNet, proposed by E. Gibson et al. [7] for volumetric, abdominal multi-organ segmentation. It comprises of a backbone encoder (series of downsampling layers used for feature extraction) followed by upsampling (via bilinear interpolation), spatial priori addition and a final convolutional layer to produce the output 3D segmentation mask. The network is pre-trained using the expanded XCAT dataset to achieve 4-fold cross-validation Dice coefficient of 0.90 for the lungs. The second sub-model is a 3D adaptation of ResNet [8], performing binary classification. Cross-entropy loss is used with the Adam optimizer in backpropagation through the model.

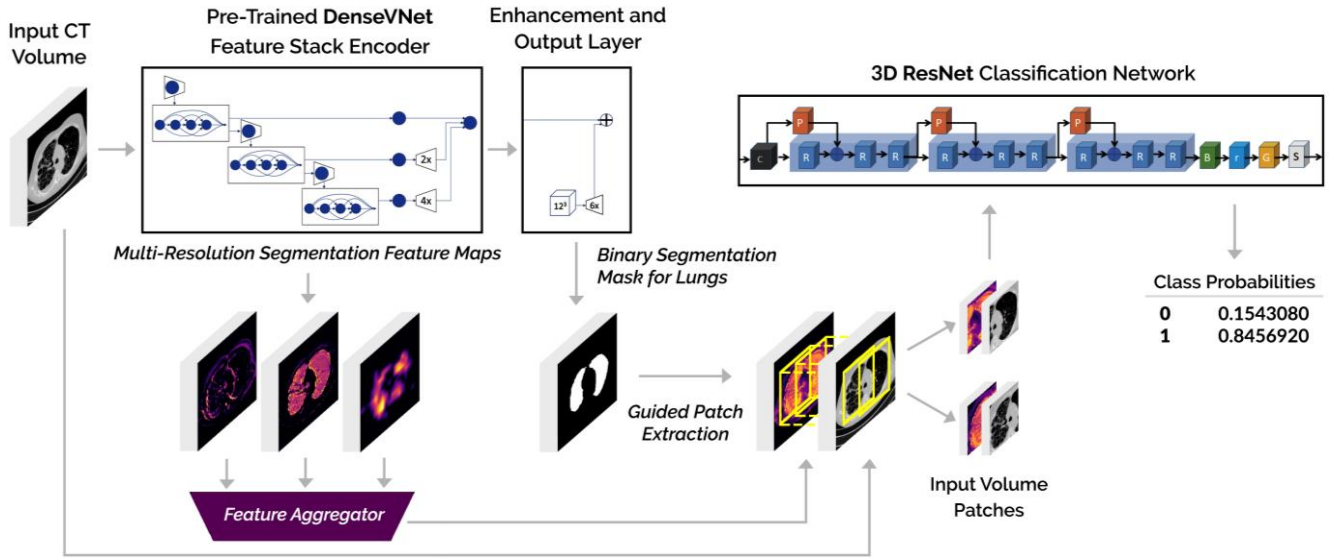

Figure 1. Dual-stage 3D CNN model architecture for reusing segmentation feature maps in binary classification. DenseVNet is used as the segmentation sub-model, taking a single channel input volume with variable dimensions. Meanwhile, the classification sub-model is a 3D ResNet with 10 residual blocks across 5 resolutions, taking a dual channel input volume patch of size [112,112,112]. Final output is a tensor with the predicted class probabilities. Both models are connected by the feature aggregator and the patch extraction algorithm, which rely on the raw multi-resolution feature maps and the output binary segmentation mask, respectively, from the DenseVNet.

At train-time, all weights for the DenseVNet are frozen. As an input CT volume passes through its cascade of dense feature stacks, segmentation feature maps are generated at multiple resolutions. From here, a small fraction is selected and passed on to the feature aggregator, which combines them to a single fused volume. Finally, the output segmentation mask is used to guide patch extraction from the 4D concatenation of the original CT volume and the volume of aggregated feature maps. The resultant multi-channel 3D volume patches are then fed to the 3D ResNet as its input for classification. During inference, 6 patches are sampled per scan and their mean predicted class probabilities are used to determine the overall prediction for the complete scan.

## 2.3 Feature Aggregation

After pre-training on the XCAT dataset, all filters along the segmentation model are tuned to detect key anatomical features that can help it distinguish between 29 unique structures pertaining to the human body. These discriminative features can range from simple edges and textures to highly complex structural concepts at the deeper levels of the encoder, as shown in Figure 2. For the DenseVNet, they exist in the form of 60 volumetric feature maps at 3 different resolutions. Although, interlaced features applicable for all 29 classes can likely provide useful peripheral information for classifying lungs, we select only 13 maps, with notable activation for either/both lungs, to ensure passing the most relevant information. For a future multi-organ approach, potentially all 60 maps can be salvaged to take maximum advantage of the system. Using the segmentation body mask generated by DenseVNet, a threshold is applied across all feature maps, to mute all activations in the external airspace outside the body. Finally, they are fused in 2 different approaches, resulting in two sub-categories of our proposed architecture: static feature aggregation (StFA) and dynamic feature aggregation (DyFA).

For StFA, we assign equal weight to each feature map, thereby numerically averaging all feature volumes along equivalent voxels. This allows us to directly control and monitor the level of supplementary semantic information entering the classification sub-model and ensure overall network stability.

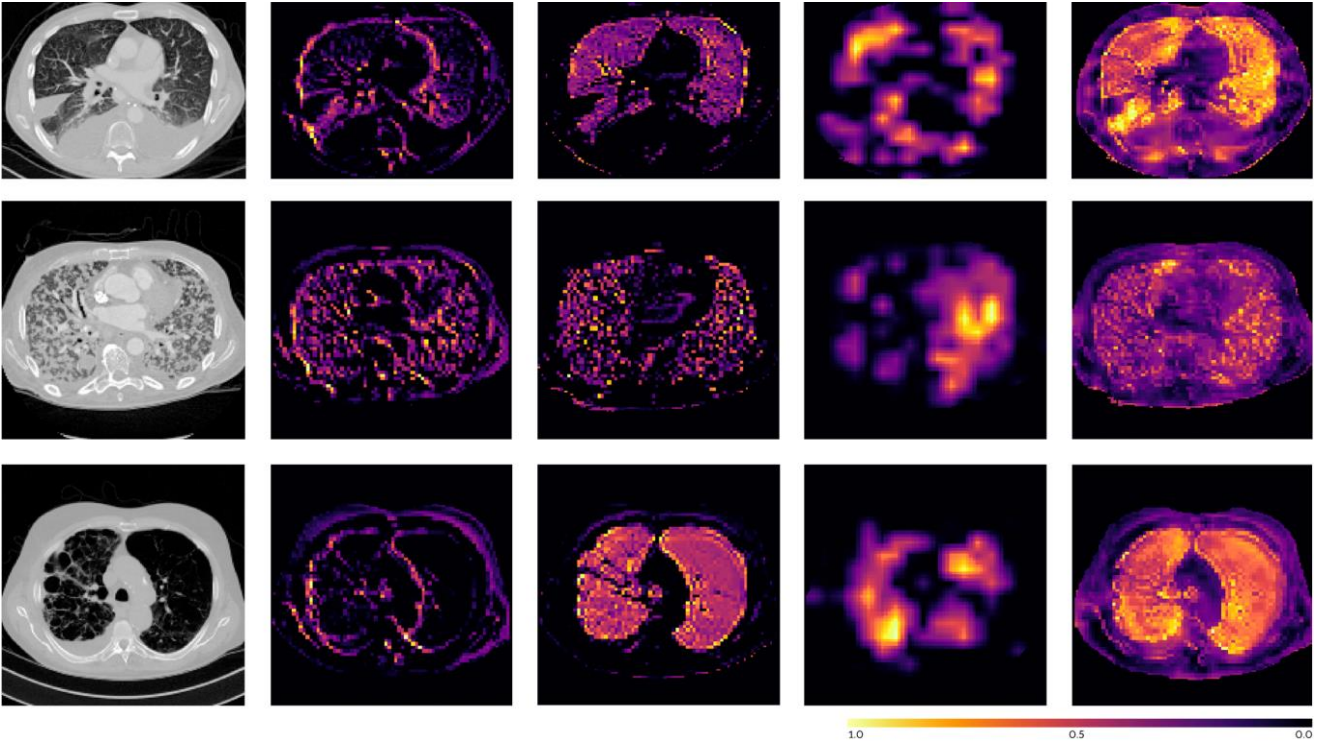

Figure 2. From left-to-right: input CT volume (axial view), 3 out of 61 segmentation feature maps extracted from the pretrained DenseVNet model at different resolutions, and their corresponding aggregated feature map using the StFA strategy, in the case of diseased lungs with atelectasis (top row), mass (middle row) and emphysema (bottom row).

For DyFA, all 13 feature maps are passed through an intermediate trainable 1×1 3D convolutional layer. As a result, each feature map receives a learnable weight (as determined by Adam optimization of the classification loss), and the final aggregated feature volume passing into the 3D ResNet remains dynamic throughout the training stage.

## 3. EXPERIMENTAL RESULTS

TensorFlow Estimator API is used for the complete implementation of the model and a single NVIDIA Titan Xp is used for train-time acceleration. The classification dataset is partitioned into stratified train/validation/test subsets of 67.5/22.5/10%, respectively. Special care is taken to ensure that the test set not only holds exclusive scans, but also exclusive patients from the other subsets. Both models (StFA, DyFA) are trained over 60 epochs using a batch size of 16 and a cyclic learning rate oscillating between $10^{-3}$–$10^{-7}$ with 0.01% exponential decay per cycle. To draw out a complete comparison, an independent 3D ResNet classification model is also trained using identical input patches, with the exception of concatenating any segmentation feature maps. Validation performance is reported separately for each of the 4 diseases (and collectively as a single positive class) vs the shared negative class, per model. Figure 3 illustrates an overall increase in performance for StFA to AUC=0.832 and DyFA to AUC=0.851, relative to an independent model (AUC=0.791), thereby verifying the additional contribution of segmentation feature maps in weakly supervised 3D classification. For the DyFA model, AUC performance for classifying individual diseases were in increasing order: nodules (0.803), pneumonia-atelectasis (0.808), mass (0.878) and emphysema (0.922). Each of these is better than the respective StFA performance, except for mass, which is slight worse (0.896). However, both of the new models outperform the baseline model in every aspect.

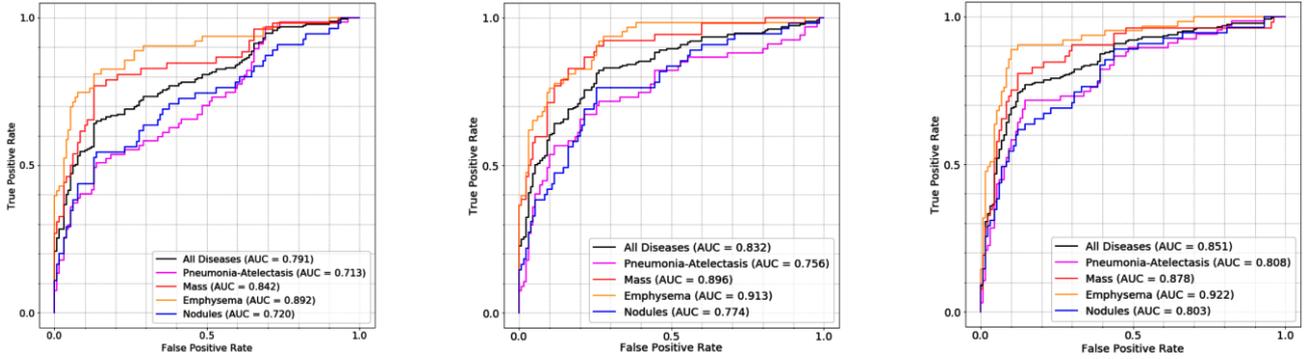

Figure 3. ROC curves for each disease class against all normal cases and all disease classes against all normal cases for the independent (left), StFA (center) and DyFA (right) models for binary lung disease classification.

## 4. LIMITATIONS AND FUTURE WORK

Preliminary trials indicate that allowing the DenseVNet to finetune any further during the classification training stage causes heavy degradation to its feature maps and overall capacity. Furthermore, over long periods of training, this tends to overwhelm the subsequent 3D ResNet, causing numerical instability along its layers. This issue remains to be investigated further. The proposed method downsamples CT scans to a moderately low resolution to alleviate computational overhead, which can potentially be detrimental for classifying focal diseases such as nodules (refer to Fig. 3). Feature map selection for StFA and DyFA strategies is largely motivated by qualitative assessment. In future work, the feature selection framework can be revised to reflect a generalized approach, adaptable for any target organ, after which, extending the proposed model to a multi-organ, multi-label classification system will become an important point of interest for further study.

## 5. CONCLUSION

To the best of our knowledge, this was the first attempt in demonstrating weakly supervised 3D classification of CT volumes to target multiple diseases in the same organ. Furthermore, we introduce a novel dual-stage convolutional neural network architecture that fully utilizes a segmentation model by leveraging its multi-resolution deep features and aggregating them in two different approaches (static, dynamic). The corresponding binary segmentation mask is then used to perform guided patch extraction from the multi-channel stack of the original CT volume and its aggregated, multi-resolution segmentation feature volumes. Finally, resultant patches are injected into the classification network as its input. We demonstrate that this alternative technique reinforces learning and improves the performance of a weakly supervised medical classifier, relative to conventional practices of directly feeding CT patches into an independent classification network.

## 6. ACKNOWLEDGEMENTS

This work was supported in part by developmental funds of the Duke Cancer Institute as part of the NIH/NCI P30 CA014236 Cancer Center Support Grant. We are grateful for GPU equipment from NVIDIA Corporation.